\documentclass[journal]{IEEEtran}
\usepackage{microtype}
\usepackage{caption}
\usepackage{graphicx}
\usepackage{booktabs} 
\usepackage{graphicx}
\usepackage{amsmath}
\usepackage{hyperref}
\usepackage{amssymb}
\usepackage{multirow}
\usepackage{subcaption}
\usepackage{booktabs}
\usepackage{colortbl}  
\usepackage{xcolor}
\usepackage{array}  
\usepackage{booktabs}

\ifCLASSOPTIONcompsoc
  \usepackage[nocompress]{cite}
\else
  \usepackage{cite}
\fi
\ifCLASSINFOpdf
\else
\fi

\begin{document}

\title{Geometry-Guided Self-Supervision for Ultra-Fine-Grained Recognition with Limited Data}

\author{Shijie Wang, Yadan Luo, Zijian Wang, Haojie Li,
        Zi Huang and Mahsa Baktashmotlagh
\IEEEcompsocitemizethanks{\IEEEcompsocthanksitem Corresponding author: Haojie Li. 
\IEEEcompsocthanksitem Shijie Wang is with College of Computer Science and Engineering, Shandong University of Science and Technology, China and with School of Information Technology and Electrical Engineering, The University of Queensland, Australia (e-mail: shijie.wang@uq.edu.au).
\IEEEcompsocthanksitem Haojie Li is with College of Computer Science and Engineering, Shandong University of Science and Technology, China (e-mail: hjli@sdust.edu.cn).
\IEEEcompsocthanksitem Yadan Luo, Zijian Wang, Zi Huang and Mahsa Baktashmotlagh are with School of Information Technology and Electrical Engineering, The University of Queensland, Australia
(e-mail: y.luo@uq.edu.au; zijian.wang@uq.edu.au, helen.huang@uq.edu.au; m.baktashmotlagh@uq.edu.au).}
\thanks{Manuscript received April 19, 2005; revised August 26, 2015.}
}

\markboth{Journal of \LaTeX\ Class Files,~Vol.~14, No.~8, August~2015}%
{Shell \MakeLowercase{\textit{et al.}}: Bare Advanced Demo of IEEEtran.cls for IEEE Computer Society Journals}

\IEEEtitleabstractindextext{%
\begin{abstract}
This paper investigates the intrinsic geometrical features of highly similar objects and introduces a general self-supervised framework called the Geometric Attribute Exploration Network (GAEor), which is designed to address the ultra-fine-grained visual categorization (Ultra-FGVC) task in data-limited scenarios.
Unlike prior work that often captures subtle yet critical distinctions, GAEor generates geometric attributes as novel alternative recognition cues. These attributes are determined by various details within the object, aligned with its geometric patterns, such as the intricate vein structures in soybean leaves.
Crucially, each category exhibits distinct geometric descriptors that serve as powerful cues, even among objects with minimal visual variation -- a factor largely overlooked in recent research. GAEor discovers these geometric attributes by first amplifying geometry-relevant details via visual feedback from a backbone network, then embedding the relative polar coordinates of these details into the final representation.
Extensive experiments demonstrate that GAEor significantly sets new state-of-the-art records in five widely-used Ultra-FGVC benchmarks. 
\end{abstract}

}

\maketitle

\IEEEdisplaynontitleabstractindextext

%
\IEEEpeerreviewmaketitle

\ifCLASSOPTIONcompsoc
\IEEEraisesectionheading{\section{Introduction}\label{sec:introduction}}
\else
\section{Introduction}
\label{sec:introduction}
\fi

%
%
%
%

\IEEEPARstart{U}{ltra}-fine-grained visual categorization (Ultra-FGVC) involves identifying objects at an exceptionally fine granularity, such as distinguishing between different soybean cultivars in smart agriculture. 
This task has attracted significant research attention due to its critical role in domain-specific applications ranging from agricultural automation to biodiversity monitoring. 
Nevertheless, Ultra-FGVC faces two fundamental challenges. First, even human experts struggle to distinguish subtle inter-class differences, such as morphological variations among cotton subspecies. Second, the extremely limited training data (\textit{e.g.}, only three samples per class in the Cotton80 and SoyLoc datasets \cite{DBLP:conf/aaai/YuZGXY20}) severely hinders recognition models from learning robust discriminative features and generalizing effectively.

\begin{figure}
    \centering
    \includegraphics[width=0.75\linewidth]{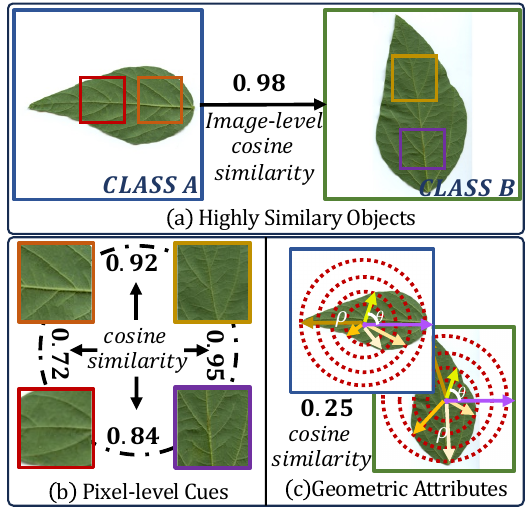}
    \caption{Pixel-level cues \textit{v.s.} Geometric attributes. Pixel-level cues are often extremely similar and challenging to distinguish, whereas exploring the intrinsic geometrical features of highly similar objects can reveal alternative discriminative patterns, namely geometric attributes. }
    \label{fig:introduction1a}
\end{figure}

Recent Ultra-FGVC research~\cite{DBLP:journals/tcsv/FangJTL24, DBLP:journals/pr/ChenJWDLWW24, DBLP:conf/ijcai/0001WG23, DBLP:conf/wacv/PanYZG23} has largely focused on data augmentation and attention mechanisms to capture discriminative details from limited data.
Although these approaches improve local detail extraction (\textit{e.g.}, soybean leaf texture), their reliance on pixel-level visual cues inherently limits discrimination of morphologically high-similarity categories. 
As shown in Fig.~\ref{fig:introduction1a}(b), two soybean cultivars exhibit nearly identical local leaf textures while possessing divergent venation topologies—structural disparities that current Ultra-FGVC works fail to detect, resulting in indistinguishable feature representations.
This limitation primarily stems from their reliance on pixel-level operations, sensitivity to geometric transformations, and inability to model structural abstractions~\cite{DBLP:journals/tog/VinkerPBBBCZS22, DBLP:conf/aaai/ChenCZT22}. 
This raises a compelling question: rather than seeking only much finer pixel-level discrepancies, can we leverage the intrinsic geometrical structure of objects to discover more robust discriminative cues? 

Interestingly, prior non-deep-learning studies \cite{DBLP:journals/tip/HuJLH12, DBLP:journals/pami/LingJ07, DBLP:journals/tip/WangG14} have extensively demonstrated that objects exhibit distinct geometric and morphological patterns, even when they appear highly similar.  
As illustrated in Fig.~\ref{fig:introduction1a}(c), the vein layout of a soybean leaf forms a distinctive geometric pattern that provides more discriminative information than purely pixel-level differences.
Therefore, our key insight is to bridge deep learning with geometric analysis: instead of only seeking "where to look" (attention), we should also learn "how to connect" (geometry). 
By guiding networks to learn characteristic geometry patterns of diverse details, we enhance their ability to achieve robust discrimination, even in scenarios where textures exhibit near-identical appearances.

\begin{figure}[t]
    \centering
    \includegraphics[width=1\linewidth]{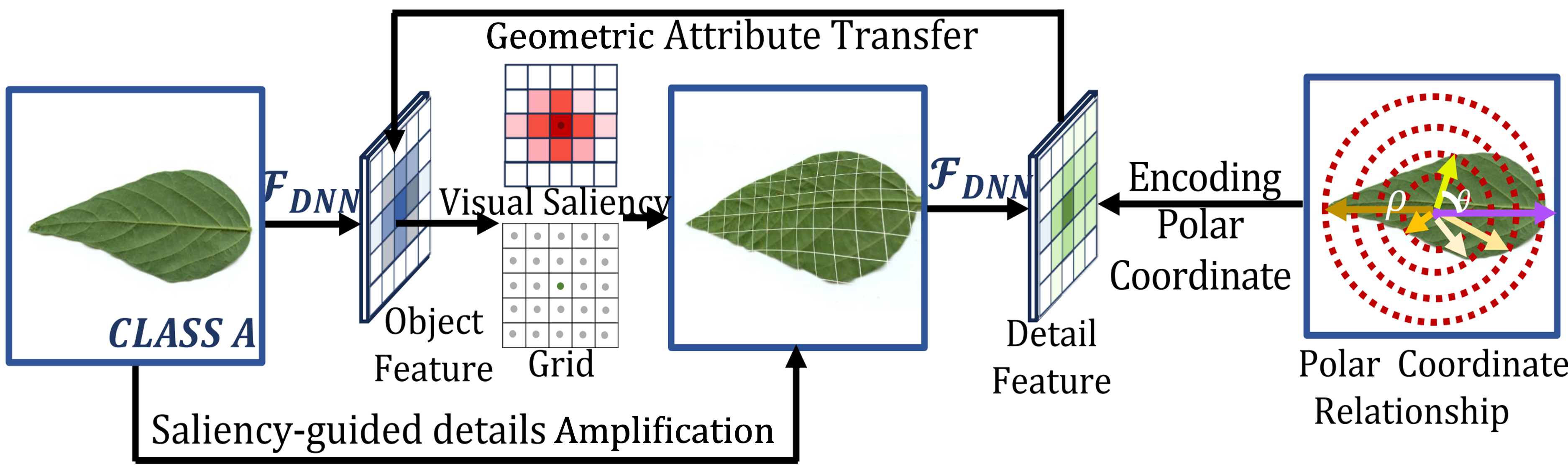}
    \caption{Discovering geometric attributes with geometry-guided self-supervision. GAEor initially amplifies task-relevant details for geometric attribute modeling and subsequently incorporates relative polar coordinates into these amplified details, ultimately generating discriminative geometric attributes.}
    \label{fig:introduction1b}
\end{figure}
Learning geometry patterns within an object from limited data poses a fundamental challenge, as explicit geometric annotations are unavailable, making it hard to establish structured supervision signals. 
To overcome this, we propose GAEor, a novel Geometric Attribute Exploration Network, which autonomously discovers geometric associations of object details through a self-supervised framework. 
As shown in Fig.~\ref{fig:introduction1b}, GAEor establishes a self-supervised paradigm for alternative discriminative cues by synergizing latent geometric reasoning with pixel-level detail refinement. 
By discovering the geometric relationships of pixel-level details, the network establishes geometric descriptors that complement pixel-level visual cues, thus uncovering novel alternative recognition cues. 
Importantly, the detail amplification process generates diverse local pattern variations during training, while integrated geometry-guided self-supervision injects additional supervisory signals, thereby jointly improving model robustness in data-limited scenarios.

Technically, 
GAEor begins by dynamically quantifying geometry-relevant details from classification representations through gradient-guided visual salience. 
This facilitates selective amplification of these details while suppressing redundant patterns in the pixel space, ensuring more precise and efficient details for geometric association modeling.
Crucially, GAEor establishes geometric associations of these amplified details through a Cartesian-to-polar coordinate transformation, where the most salient detail is designated as the coordinate origin. It then computes angular-radial displacements for all amplified details, treating them as self-supervised geometry invariants of the input object.
These invariants are further incorporated into the representations of amplified details through a polar coordinate-based prediction objective, enabling the encoding of geometric associations among pixel-level details and acquiring geometric attributes as alternative discriminative cues.
Finally, to adapt geometric attribute learning for highly similar object recognition, a knowledge distillation module progressively transfers geometric attributes from the self-supervised pathway to the classification branch, enabling efficient inference without additional computational overhead.

Our major contribution is four-fold:
\begin{itemize}
    \item 
     To the best of our knowledge, we are the first to shift the focus from capturing extremely subtle discrepancies to investigating the geometric structures of pixel-level details within objects, discovering novel alternative recognition cues for Ultra-FGVC.
    \item 
    We introduce GAEor, a novel geometry-guided self-supervised framework that autonomously discovers geometric associations of object details, requiring only category labels as supervision.
    \item 
    GAEor leverages gradient-guided saliency to dynamically amplify discriminative details, then encodes their spatial configurations into stable geometric invariants through Cartesian-to-polar coordinate transformations, enabling annotation-efficient reasoning of geometric attributes.
    \item 
    Our comprehensive experimental analysis, along with achieving new state-of-the-art performance on five widely-used Ultra-FGVC benchmarks, demonstrates the effectiveness of our approach.
\end{itemize} 

\begin{figure*}[!t]
    \centering
    \includegraphics[width=1\linewidth]{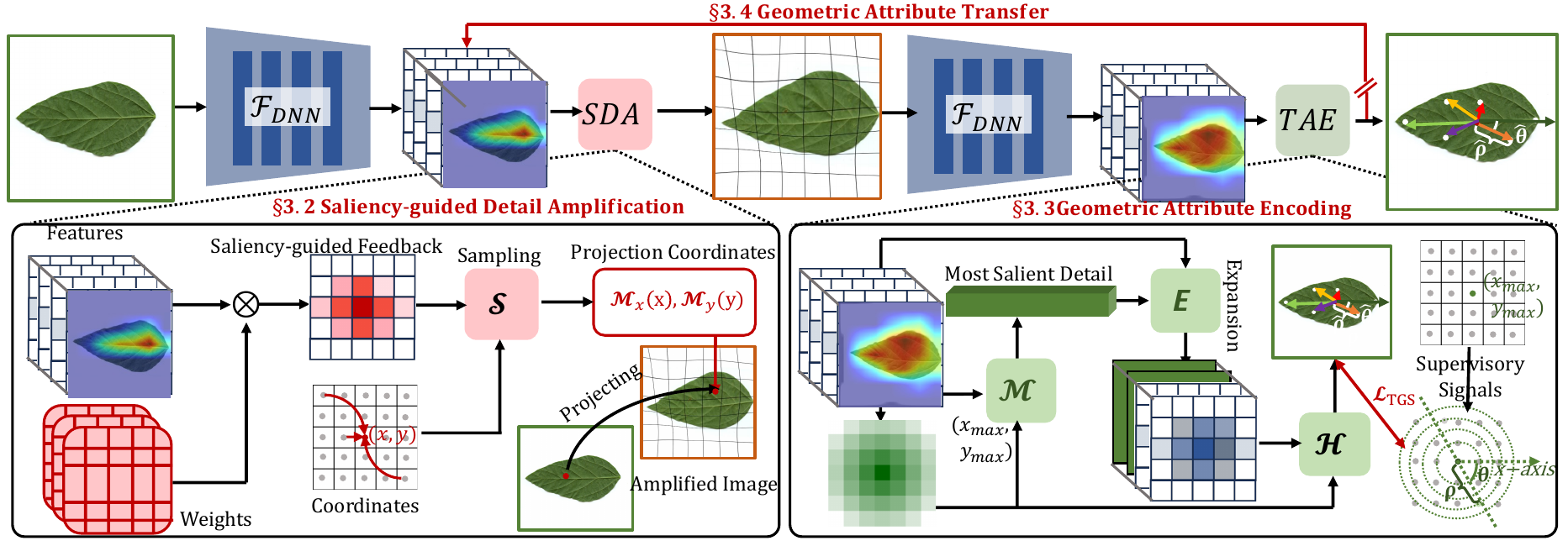}
    \caption{Detailed illustration of {
   \bf Geometric Attribution Exploration} network.
Our framework is composed of three key components: a Saliency-guided Detail Amplification (SDA) module, a Geometric Attribute Encoding (GAE) module, and a Geometric Attribute Transfer (GAT) module.
The SDA module dynamically identifies discriminative details from classification representations using gradient-guided visual saliency and amplifies these details in the pixel space through inhomogeneous sampling.
The GAE module encodes spatial geometry into the amplified details by introducing self-supervised signals derived from Cartesian-to-polar coordinate transformations, thereby capturing geometric attributes.
Finally, the GAT module facilitates efficient recognition of visually similar objects by progressively transferring learned geometric attributes from the self-supervised branch to the classification branch.
The symbols used in this figure can be found in \S\ref{sec3}. }
    \label{fig:method}
\end{figure*}

\section{Related Work}

\textbf{Ultra-fine-grained visual categorization.}
Ultra-Fine-Grained Visual Categorization (Ultra-FGVC), initially formalized by Yu et al.~\cite{DBLP:conf/iccv/YuZ0YX21}, seeks to distinguish highly similar subcategories within fine-grained classes (e.g., soybean cultivars differentiated by $\le$ 3 genetic markers). 
Their work introduced diverse leaf-based benchmarks, such as SoyAgeing (representing phenotypic aging stages) and SoyGene (capturing genotype-driven variations), where subcategory labels are derived from seed genetic repository IDs, ensuring biological precision in categorization.
Existing approaches \cite{DBLP:conf/aaai/YuZGXY20, DBLP:conf/iccv/YuZ0YX21, DBLP:conf/cvpr/LiuCJQ0P24, DBLP:journals/tcsv/FangJTL24, DBLP:journals/pr/YuZG22, rios2025cross, DBLP:conf/icmlc2/HuangZWQ24, DBLP:conf/dicta/AkpudoYZG23} tackle extreme inter-class similarity by uncovering highly subtle visual discrepancies. These methods can be categorized into two distinct groups as follows.

The first group, \textbf{attention-based manner} \cite{DBLP:journals/pr/YuZGX21, DBLP:journals/pr/YuZG22, DBLP:conf/bmvc/WangY021}, focuses on capturing discriminative details from highly similar objects by leveraging attention mechanisms, similar to the approaches used in fine-grained image classification tasks \cite{DBLP:conf/aaai/HeCLKYBW22, DBLP:journals/tip/SunHXP24, DBLP:journals/pami/WeiSAWPTYB22, sun2023fine}. 
For instance, SPARE \cite{DBLP:journals/pr/YuZG22} segments object parts using only image-level category labels, thereby generating discriminative part-based feature representations.
UFG-NCD \cite{DBLP:conf/cvpr/LiuCJQ0P24} extracts and utilizes discriminative features from local regions by exploring semantic alignment between regions along the channel direction.
The second group, \textbf{data augmentation-based methods}, aims to learn discriminative feature representations through either image or feature augmentation techniques \cite{DBLP:conf/ijcai/0001WG23, DBLP:journals/pr/ChenJWDLWW24, DBLP:journals/tcsv/FangJTL24, DBLP:journals/pr/YuWZG23}. 
For image augmentation, works such as CLE-ViT \cite{DBLP:conf/ijcai/0001WG23} and FDCL-DA \cite{DBLP:journals/pr/ChenJWDLWW24} divide images into grids of square patches and apply augmentations, such as erasing or shuffling specific patches, which helps generate patch-level feature representations. 
For feature augmentation, Mix-ViT \cite{DBLP:journals/pr/YuWZG23} introduces a contrastive token substitution framework, where high-level vision transformer tokens are selectively swapped based on attention saliency, thus discovering discriminative details by focusing on the most informative regions.

While they primarily focus on pixel-level visual cues through attention mechanism or data augmentation strategies, they inherently overlook local details that frequently recur across other highly similar samples, leading to insufficient discriminative signals. 
To overcome this limitation, we propose a novel Geometry-Guided Self-Supervision paradigm that explores geometric attributes by embedding geometric associations within object details. 
Our approach uncovers alternative discriminative cues, offering a more robust and effective means of distinguishing highly similar samples.

\textbf{Polar representation.} Polar representation \cite{DBLP:journals/rti/DenzlerN99} has previously been utilized to describe polygons in tasks such as instance segmentation \cite{DBLP:journals/pami/XieWDZL22}, pose estimation \cite{DBLP:journals/tip/LiWZ23}, time series forecasting \cite{DBLP:conf/aaai/LiXA24} and 3D object detection \cite{DBLP:conf/iccv/NieXWYXZHMW023}. 
PolarMask \cite{DBLP:journals/pami/XieWDZL22} specifies 36 predetermined orientations and estimates the lengths from a central point towards these 36 directions, thereby constructing the polygon contour.
PolarPose \cite{DBLP:journals/tip/LiWZ23} simplifies the task of regressing 2D offsets in Cartesian coordinates by converting it to associated tasks in Polar coordinates, which are generally easier to optimize.
Recent work in large language models \cite{DBLP:conf/nips/Diego-SimondCLK24} leverages polar representations to capture syntactic relationships by interpreting both the distance and direction between word embeddings.
Inspired by these works, we integrate polar coordinates into the visual cues of an object to establish its geometric associations, thereby generating alternative discriminative cues for identifying visually high-similarity objects.

\textbf{Self-supervised Learning.} 
Self-supervised learning \cite{DBLP:journals/pami/GuiCZCSLT24, DBLP:journals/tkde/LiuJPZZXY23} has emerged as a transformative paradigm in visual representation learning, leveraging the inherent structure of unlabeled data to reduce dependence on large-scale annotated datasets. 
It effectively improves model efficiency and generalization by enabling neural networks to learn transferable and discriminative features through carefully designed pretext tasks. 
Recent advances have led to significant progress across a wide range of computer vision tasks, including image classification \cite{DBLP:conf/cvpr/MisraM20, DBLP:conf/cvpr/EricssonGH21}, semantic segmentation \cite{DBLP:conf/cvpr/WangZKSC20, DBLP:conf/cvpr/ChenYLX22}, 3D representation learning \cite{DBLP:conf/iccv/HuangXZZ21, DBLP:conf/cvpr/Wang00TPZJ24}, vision-language alignment \cite{DBLP:conf/cvpr/YangDTXCCZCH22, DBLP:journals/tmm/ZhuangYDQH24}, and domain-specific applications such as autonomous driving \cite{DBLP:journals/ral/KahnAL21, DBLP:conf/cvpr/SautierPGBBM22} and medical imaging \cite{DBLP:conf/iccv/AziziMRBFDLKKCN21, DBLP:conf/nips/TalebLDSGBL20}. 
Moreover, self-supervised methods have been employed to explore complex intra-object structures, such as part-level alignment \cite{DBLP:conf/aaai/WangWLO22, DBLP:conf/eccv/CheFK24} and relational modeling among object components \cite{DBLP:conf/cvpr/ZieglerA22, DBLP:conf/nips/PatacchiolaS20}.
Inspired by these advances, we compute the polar coordinates of object details from their Cartesian positions and treat them as instance-level supervisory signals to optimize the retrieval model for geometric attribute modeling.

\section{Methodology}
\label{sec3}
In this section, we introduce GAEor, a novel framework whose overall architecture is depicted in Fig.~\ref{fig:method}, aiming to explore geometric attributes within objects rather than merely capturing hard-to-identify pixel-level discrepancies.
GAEor first leverages saliency-guided feedback from the classification branch (\S\ref{Sec3.1}) to selectively amplify discriminative details (\S\ref{sec3.2}). 
It then predicts the relative polar coordinates of the amplified details supervised by Cartesian-to-polar transformations, to encode their spatial geometry and thereby acquire geometric attributes (\S\ref{sec3.3}). 
Finally, these geometric attributes are distilled into the classification branch to collaboratively enhance the discrimination of highly similar objects (\S\ref{3.4}).


\subsection{Classification Branch}
\label{Sec3.1}
The classification branch is designed to extract object representations enriched with geometric attributes using a backbone network trained in a self-supervised manner, subsequently generating category predictions.
Formally, given a highly similar object $\mathbf{I} \in\mathbb{R}^{\mathrm{3\times H_I \times W_I}}$ and its ground truth one-hot label $\mathbf{l}$, let $\mathbf{F} \in \mathbb{R}^{\mathrm{C \times H \times W}}$ represent the $\mathrm{C}$-dimensional feature map with spatial dimensions $H \times W$, encoded by the backbone network, such that $\mathbf{F} = \mathcal{F}_{\mathrm{DNN}}(\mathbf{I})$. The probability vector for the category prediction is denoted as $\mathbf{y}(\mathbf{I})$. The loss function of the classification branch, denoted as $\mathcal{L}_{\mathrm{CLS}}$, is defined as:  
\begin{equation}  
    \mathcal{L}_{\mathrm{CLS}} = -\sum_{\mathbf{I} \in \mathcal{I}} \mathbf{l} \cdot \log \mathbf{y}(\mathbf{I}),  
\end{equation}  
where $\mathcal{I}$ represents the set of training images. 
It should be clarified that GAEor incurs no additional computational overhead during evaluation, as only the classification branch is utilized in this stage.

\begin{figure}
    \centering
    \includegraphics[width=0.7\linewidth]{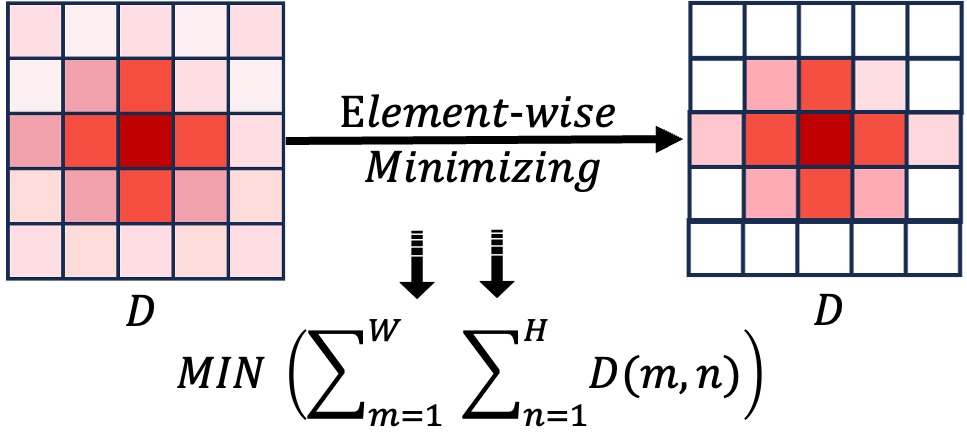}
    \caption{A regularization constraint is used to suppress the geometry-irrelevant activation. }
    \label{fig:reg_loss}
\end{figure}

\subsection{Saliency-guided Detail Amplification}
\label{sec3.2}
Recognizing geometric descriptors within highly similar objects is essential for extracting alternative discriminative cues in ultra-fine-grained visual categorization. 
However, such descriptors often appear as subtle, semantically ambiguous patterns that challenge even expert annotations. 
To address this, we introduce a Saliency-Guided Detail Amplification (SDA) module composed of two stages: saliency-guided detail feedback and geometry-relevant detail amplification. 
In the first stage, SDA dynamically identifies geometry-relevant regions by computing gradient-based visual saliency through backward gradient propagation from the classification and geometric modeling branches. 
In the second stage, it amplifies these geometry-relevant details in the pixel space via inhomogeneous sampling, strategically emphasizing high-saliency regions while maintaining spatial coherence.
In this way, SDA can enhance the model’s sensitivity to subtle geometry-relevant details critical for geometric modeling.

\textbf{Saliency-guided detail feedback.} 
We design a lightweight generator $\mathcal{G}(\cdot)$ that projects the representation $\mathbf{F}$ into a feedback map $\mathbf{D} \in\mathbb{R}^{\mathrm{H \times W}}$, specifying the location and scale of geometry-relevant details:
\begin{equation}  
    \mathbf{D} = \mathcal{G}(\mathbf{F}), 
\end{equation}  
where $\mathcal{G}$ is implemented as a convolutional layer with a kernel size of 1 followed by a sigmoid function. 
Furthermore, the feedback map should focus exclusively on geometry-relevant details that are essential for accurate geometric attribute modeling.
As shown in Fig.~\ref{fig:reg_loss}, we implement a regularization constraint to constrain the activation size within the feedback map, thereby ensuring that irrelevant details are excluded: 
\begin{equation}
    \mathcal{L}_{\mathrm{REG}} = \frac{1}{\mathrm{W \times H}} \sum_{\mathrm{m = 1}}^\mathrm{W} \sum_{\mathrm{n = 1}}^\mathrm{H} \mathbf{D}(\mathrm{m}, \mathrm{n}).
\end{equation}

Notably, while the current implementation does not explicitly incorporate gradients, the training process implicitly aligns the saliency-guided feedback \( \mathbf{D} \) with gradient-based saliency patterns. 
Since GAEor is an end-to-end framework, the gradients from the classification loss, regularization constraint, and subsequent geometry-guided self-supervised loss (\S~\ref{sec3.3}) are jointly backpropagated to update the generator \( \mathcal{G}(\cdot) \), thereby guiding it to focus on regions where fine-grained details critically influence geometric attribute modeling.
As a result, \( \mathbf{D} \) functions as a self-calibrating importance estimator that adaptively emphasizes geometry-relevant details that constitute key geometric configurations.

\textbf{Geometry-relevant detail amplification.} 
We leverage feedback on the location of geometry-relevant details to learn a mapping function that determines the position of pixels from the original input image in the transformed image, effectively rearranging object content through an inhomogeneous transformation to amplify these details.
For example, given the coordinate $(\mathrm{x, y})$ in the transformed image $\mathbf{I_D}$, the mapping coordinate $(\mathrm{\mathcal{M}_{x}}, \mathrm{\mathcal{M}_{y}})$ in the original input image $\mathbf{I}$ can be calculated as follows:
\begin{equation}
\begin{split}
    &\mathcal{M}_\mathrm{x}(\mathrm{x}) = \\ &\frac{\sum_{\mathrm{w=1}}^{\mathrm{W_I}}\sum_{\mathrm{h=1}}^{\mathrm{H_I}} \mathbf{D}(\mathrm{w}, \mathrm{h}) \cdot \mathcal{K}<(\frac{\mathrm{x}}{\mathrm{W_I}},\frac{\mathrm{y}}{\mathrm{H_I}}),(\frac{\mathrm{w}}{\mathrm{W_I}},\frac{\mathrm{h}}{\mathrm{H_I}})>\cdot \frac{\mathrm{x}}{\mathrm{W_I}}}{\sum_{\mathrm{w=1}}^{\mathrm{W_I}}\sum_{\mathrm{h=1}}^{\mathrm{H_I}} \mathbf{D}(\mathrm{w}, \mathrm{h}) \cdot \mathcal{K}<(\frac{\mathrm{x}}{\mathrm{W_I}},\frac{\mathrm{y}}{\mathrm{H_I}}),(\frac{\mathrm{w}}{\mathrm{W_I}},\frac{\mathrm{h}}{\mathrm{H_I}})>},
    \\
   & \mathcal{M}_\mathrm{y}(\mathrm{y}) =\\ &\frac{\sum_{\mathrm{w=1}}^{\mathrm{W_I}}\sum_{\mathrm{h=1}}^{\mathrm{H_I}} \mathbf{D}(\mathrm{w}, \mathrm{h}) \cdot \mathcal{K}<(\frac{\mathrm{x}}{\mathrm{W_I}},\frac{\mathrm{y}}{\mathrm{H_I}}),(\frac{\mathrm{w}}{\mathrm{W_I}},\frac{\mathrm{h}}{\mathrm{H_I}})>\cdot \frac{\mathrm{y}}{\mathrm{H_I}}}{\sum_{\mathrm{w=1}}^{\mathrm{W_I}}\sum_{\mathrm{h=1}}^{\mathrm{H_I}} \mathbf{D}(\mathrm{w}, \mathrm{h}) \cdot \mathcal{K}<(\frac{\mathrm{x}}{\mathrm{W_I}},\frac{\mathrm{y}}{\mathrm{H_I}}),(\frac{\mathrm{w}}{\mathrm{W_I}},\frac{\mathrm{h}}{\mathrm{H_I}})>},
    \end{split}
    \label{eq4}
\end{equation}
where $\mathcal{K}<,>$ represents the Gaussian distance kernel, which acts as a regularizer to preserve the relative positional relationships between pixels from the global perspective, thus preventing structure distortion during the projection process. 

Then, we use the differentiable bi-linear sampling mechanism, which linearly interpolates the values of the 4-neighbors (top-left, top-right, bottom-left, bottom-right) of $(\mathrm{\mathcal{M}_{x}(x)}, \mathrm{\mathcal{M}_{y}(y)})$, denoted by $\mathcal{N} (\mathrm{\mathcal{M}_{x}(x)}, \mathrm{\mathcal{M}_{y}(y)})$  to approximate the final output, denoted by 
\begin{equation}
    \mathbf{I_D}(\mathrm{x,y}) = \sum_{\mathrm{(i, j) \in \mathcal{N} (\mathrm{\mathcal{M}_{x}(x)}, \mathrm{\mathcal{M}_{y}(y)}) }} \mathrm{W_D} \cdot \mathbf{I}(\mathrm{i, j}),
    \label{eq5}
\end{equation}
where $\mathrm{W_D}$ is the bi-linear kernel weights estimated by the distance between the mapping point and its neighbors.

Equations~(\ref{eq4})-(\ref{eq5}) unveil an adaptive sampling mechanism that iteratively reinforces high-significance pixels—identified via feedback-guided importance metrics—through multiple sampling passes, while progressively suppressing low-significance regions via sparse sampling. 
This process introduces global contextual reasoning during pixel selection for each spatial location in $I_D$. 
As shown in Fig.~\ref{fig:zoom}, it enables controlled geometric perturbations (e.g., subtle positional shifts or scale variations) to enhance geometry-relevant details and preserve geometric integrity across the entire object.
Therefore, our SDA constitutes a controlled deformation paradigm that amplifies geometry-relevant details while accommodating minor geometric adjustments, thereby enhancing the discriminability of geometry-relevant details without compromising geometric semantics.

\subsection{Geometric Attribute Encoding}
\label{sec3.3}

\begin{figure}
    \centering
    \includegraphics[width=1\linewidth]{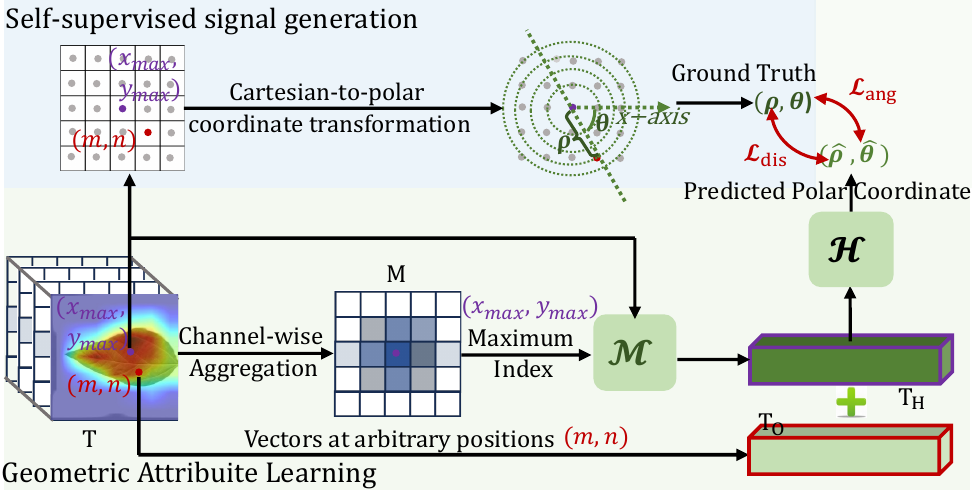}
    \caption{An overall of the process of Geometric Attribute Encoding. }
    \label{fig:tae_loss}
\end{figure}

While amplification effectively highlights details relevant to geometric attribute modeling, transforming these details into alternative discriminative cues further requires embedding their meaningful spatial relationships. 
To address this, we propose a self-supervised Geometric Attribute Encoding (GAE) module that systematically embeds spatial geometry via encoding relative polar coordinates. 
As shown in Fig.~\ref{fig:tae_loss}, GAE operates in two steps: (1) generating instance-level polar coordinate supervision signals via a Cartesian-to-polar coordinate transformation, and (2) using these signals to establish geometric associations between the learned embedding of amplified details, thereby obtaining geometric attributes. 

\textbf{Self-supervised signal generation.} 
Given the transformed image $\mathbf{I_D}$, we input it into the shared backbone network (previously used in the classification branch) to extract amplified detail representation $\mathbf{T} \in \mathbb{R}^{C\times H\times W} $.
Each spatial vector in $\mathbf{T}$ encodes semantic information of the corresponding receptive field in $\mathbf{I_D}$.
To establish self-supervision, we transform geometric associations from Cartesian to polar coordinates, deriving supervisory signals based on the spatial distribution of vectors in $\mathbf{T}$. These polar coordinates effectively capture spatial relationships among geometry-relevant details, serving as a structural prior for geometric attribute modeling.

Specifically, we first transform the detail representation $\mathbf{T}$ into a pattern map $\mathbf{M} \in\mathbb{R}^{\mathrm{H\times W}}$ by aggregating channel-wise responses through summation, and utilize this pattern map to locate the highest-activation vector $ \mathbf{T_H} = \mathbf{T(\mathrm{x_{max}, y_{max}})}$ from $\mathbf{T}$, where
\begin{equation}
  \left(\mathrm{x_{max}, y_{max}}\right)=\arg\max_{\mathrm{1\le i \le W, 1\le j \le H}}\mathbf{M}(i,j).
\end{equation}
Then, given $\mathbf{T_H}$ with spatial coordinates $(\mathrm{x_{max}, y_{max}})$ and its horizontal direction, we can express the polar coordinates of a region $\mathbf{T_O}$ located at spatial coordinates $(\mathrm{m, n})$ as: 
\begin{equation}
    \begin{split}
        \rho_{(\mathrm{m, n})} &= \sqrt{(\mathrm{x_{max}} - \mathrm{m})^2 + (\mathrm{y_{max}} - \mathrm{n})^2}/ \sqrt{\mathrm{W^2}+\mathrm{H^2}}, \\
        \theta_{(\mathrm{m, n})} &= ( \mathrm{atan2}{\left(\mathrm{y_{max}} - \mathrm{n},\mathrm{x_{max}} - \mathrm{m}\right)}+\pi)/ 2\pi.
    \end{split}
\end{equation}
where $\rho_{(\mathrm{m, n})} \in [0,1]$ represents the normalized radial distance between $\mathbf{T_H}$ and $\mathbf{T_O}$, while $\theta_{(\mathrm{m, n})}\in [0,1]$ denotes the normalized polar angle of $\mathbf{T_O} $ relative to the horizontal direction defined by $\mathbf{T_H}$.
Therefore, $(\rho_{(\mathrm{m, n})}, \theta_{(\mathrm{m, n})})$ are treated as geometry-guided displacements of pixel-level details, serving as self-supervised geometric invariants of the input object.



\textbf{Geometric attribute learning.} 
We obtain geometric attributes by predicting polar coordinates through fused features of the reference vector $\mathbf{T_H}$ and other vectors $\mathbf{T_O}$:
\begin{equation}
(\hat{\rho}_{(\mathrm{m, n})}, \hat{\theta}_{(\mathrm{m, n})}) = \mathcal{H}(\mathbf{T_H}||\mathbf{T_O}),   
\end{equation}
where $\mathcal{H}(\cdot)$ denotes the polar coordinate head implemented as a fully connected layer with ReLU activation. $(\hat{\rho}{(m, n)}, \hat{\theta}{(m, n)})$ denote the predicted radial distance and polar angle from the spatial position $(m, n)$ to a reference position $(x_{max}, y_{max})$.


To precisely model geometric attributes, we introduce a polar coordinate-based prediction objective that imposes radial distance and polar angle constraints, which supervise the representation learning process using generated geometry-guided displacements. 
Accordingly, the radial distance constraint applied to all vectors within $\mathbf{I_D}$ is formulated as follows:
\begin{equation}
    \mathcal{L}_{\mathrm{dis}} = \frac{\sum_{\mathrm{1\le m \le W},\mathrm{1\le n \le H}} \mathbf{1_{(\mathbf{M}{(\mathrm{m,n})} > \mu_{\mathbf{M}})}}\cdot||\hat{\rho}_{(\mathrm{m, n})}- \rho_{(\mathrm{m, n})}||}{\sum_{\mathrm{1\le m \le W},\mathrm{1\le n \le H}} \mathbf{1_{(\mathbf{M}{(\mathrm{m,n})} > \mu_{\mathbf{M}})}}},
    \label{eq9}
\end{equation}
where $||\cdot||$ denotes the Frobenius norm, and $\mu_{\mathbf{M}}$ represents the spatial average of the pattern map $\mathbf{M}$, which is used to implement hard attention for selecting geometry-relevant details.

Additionally, a polar angle constraint is imposed to ensure that the geometric attributes remain rotation-invariant and robust to intra-class appearance variations. 
This is implemented by computing the standard deviation of the differences between the predicted and ground-truth polar angles:
\begin{equation}
    \mathcal{L}_{\mathrm{ang}} = \frac{\sum_{\mathrm{1\le m \le W},\mathrm{1\le n \le H}} \mathbf{1_{(\mathbf{M}{(\mathrm{m,n})} > \mu_{\mathbf{M}})}} \cdot||\Delta_{\theta_{\mathrm{m,n}}}-\overline{\Delta}_{\theta}||}{\sum_{\mathrm{1\le m \le W},\mathrm{1\le n \le H}} \mathbf{1_{(\mathbf{M}{(\mathrm{m,n})} > \mu_{\mathbf{M}})}}}. 
    \label{eq10}
\end{equation}
Here, $\Delta_{\theta_{\mathrm{m,n}}} = ||\hat{\theta}_{(\mathrm{m, n})} - \theta_{(\mathrm{m, n})}||$ represents the discrepancy between the predicted and ground-truth polar angles. And $\overline{\Delta}_{\theta}$ denotes the weighted mean discrepancy across spatial locations between the predicted and ground-truth polar angles, emphasizing geometry-relevant details with higher saliency scores. It is then computed as:
\begin{equation}
\overline{\Delta}_{\theta} = 
\frac{\sum_{\mathrm{1\le m \le W},\mathrm{1\le n \le H}} \mathbf{1}_{(\mathbf{M}{(\mathrm{m,n})} > \mu_{\mathbf{M}})} \cdot\Delta_{\theta_{\mathrm{m,n}}}}{\sum_{\mathrm{1\le m \le W},\mathrm{1\le n \le H}} \mathbf{1_{(\mathbf{M}{(\mathrm{m,n})} > \mu_{\mathbf{M}})}}}.
\end{equation}


Therefore, the overall constraint of geometric attribute encoding is defined as
\begin{equation}
    \mathcal{L}_{\mathrm{GAE}} = \mathcal{L}_{\mathrm{dis}}+\mathcal{L}_{\mathrm{ang}}.
\end{equation}
By optimizing $\mathcal{L}_{\mathrm{GAE}}$ during training, our GAEor establishes geometric associations between geometry-relevant details, acquiring alternative discriminative cues that exhibit inherent rotation invariance, thus maintaining consistent representational capability regardless of object orientation.

\subsection{Geometric Attribute Transfer} 
\label{3.4}
To enable robust recognition of highly similar objects through geometric attribute learning, we design a knowledge distillation framework that progressively transfers discriminative geometric attributes from the self-supervised learning pathway to the classification branch. This strategy preserves geometric attribute awareness while eliminating the need for complex geometry-relevant detail amplification and redundant feature extraction during inference, thereby ensuring computational efficiency.

To enforce consistency between the learned geometric attributes and the classification-oriented feature representations derived directly from raw images, we propose a geometric attribute transfer constraint based on cross-pathway knowledge alignment:
\begin{equation}  
    \mathcal{L}_{\mathrm{GAT}} = ||\mathrm{g(\mathbf{T})-g(\mathbf{F})}||,  
\end{equation}  
where $\mathrm{g(\cdot)}$ denotes the global average pooling operation.

\subsection{Overall Optimization}
The total loss $\mathcal{L}$ of GAEor is defined as
\begin{equation}
    \mathcal{L} = \mathcal{L}_{\mathrm{CLS}} + \alpha \mathcal{L}_{\mathrm{REG}} + \beta \mathcal{L}_{\mathrm{GAE}} + \gamma \mathcal{L}_{\mathrm{GAT}}
    \label{eq13}
\end{equation}
where $\alpha$, $\beta$, and $\gamma$ are the hyper-parameters to balance the contributions of the individual loss item.


\begin{table}[!t]
    \centering
     \setlength{\tabcolsep}{8.5pt}

    \caption{ Ablation study of the proposed Saliency-guided Detail Amplification (SDA) and Geometric Attribute Encoding (GAE) branches, using the Cotton80 and SoyLoc datasets, respectively.}
    \begin{tabular}{c c c c c c}
    	\toprule[1pt]
       \multirow{2}{*}{Baseline} & \multirow{2}{*}{SDA}& \multirow{2}{*}{GAE} & \multirow{2}{*}{GAT}& \multicolumn{2}{c}{Accuracy} \\
       \cline{5-6}
       &&&&Cotton80 & SoyLoc \\
       \toprule[0.7pt]
       \checkmark&&&&61.3\% & 51.8\% \\
       \checkmark& \checkmark &&&63.2\% & 54.8\% \\
        \checkmark&  &\checkmark&&64.3\% & 55.7\% \\
       \checkmark&&\checkmark&\checkmark&68.4\% &58.3\% \\
       \rowcolor{gray!30}
       \checkmark&\checkmark&\checkmark&\checkmark&\textbf{71.3\%} &\textbf{62.5\%} \\

		\toprule[1pt]
	
    \end{tabular}
    \label{tab:ae}
\end{table}
\begin{table}
    \centering
    \caption{ Ablation study of using different combinations of constraints across Cotton80 and SoyLoc datasets, respectively.}
    \begin{tabular}{c c c c c c  c}
    	\toprule[1pt]
       \multirow{2}{*}{$\mathcal{L}_{\mathrm{CLS}}$} & \multirow{2}{*}{$\mathcal{L}_{\mathrm{REG}}$}& 
       \multirow{2}{*}{$\mathcal{L}_{\mathrm{GAT}}$} & \multirow{2}{*}{$\mathcal{L}_{\mathrm{dis}}$}& \multirow{2}{*}{$\mathcal{L}_{\mathrm{ang}}$}& \multicolumn{2}{c}{Accuracy} \\
       \cline{6-7}
       &&&&&Cotton80 & SoyLoc \\
       \toprule[0.7pt]
       \checkmark&&&&&62.8\% & 52.7\% \\
       \checkmark& \checkmark& & &&63.2\% & 54.8\% \\
        \checkmark&\checkmark  &\checkmark  &\checkmark &&66.8\% &  57.4\%\\
        \rowcolor{gray!30}
       \checkmark&\checkmark&\checkmark  &\checkmark&\checkmark&\textbf{71.3\%} &\textbf{62.5\%} \\

		\toprule[1pt]
	
    \end{tabular}
    
    \label{tab:ael}
\end{table}

\section{Experiments}
\subsection{Experimental Settings}
\textbf{Datasets.} 
GAEor is evaluated on five datasets \cite{DBLP:conf/aaai/YuZGXY20}, specifically, Cotton80, SoyLoc, SoyGene, SoyAgeing, and SoyGlobal.
The Cotton80 dataset comprises 80 cotton species, totaling 480 images, split into 240 images for training and 240 for testing.
The SoyLoc dataset encompasses 200 soybean species, consisting of 1,200 images, divided into 600 for training and 600 for testing.
The SoyAgeing dataset consists of 198 soybean species, with a total of 9,900 images, evenly divided into 4,950 for training and 4,950 for testing.
The SoyGlobal dataset contains 1,938 soybean species, totaling 11,628 images, split into 5,814 for training and 5,814 for testing.
The SoyGene dataset includes 1,110 soybean species, comprising 23,906 images, divided into 12,763 for training and 11,143 for testing.
\textit{These datasets suffer from limited training samples per subcategory, often consisting of only a single-digit number of training images.}

\textbf{Implementation Details.}
We adopt the Swin Transformer Base \cite{DBLP:conf/iccv/LiuL00W0LG21} as our backbone network, initializing it with pre-trained parameters from ImageNet21k \cite{DBLP:conf/cvpr/DengDSLL009}. 
The input raw images are resized to $512\times 512$ and cropped into $448 \times 448$. We train our model using Stochastic Gradient Descent (SGD) optimizer with weight decay of 0.0001, momentum of 0.9, and batch size of 16. We adopt the commonly used data augmentation techniques, \textit{i.e.}, random cropping and erasing, left-right flipping, and color jittering for robust feature representations. Our model is relatively lightweight and is trained end-to-end on four NVIDIA 2080Ti GPUs for acceleration. 
The initial learning rate is set to $ 10^{-5} $, with exponential decay of 0.9 after every 5 epochs. The total number of training epochs is set to 200.

\begin{table*}
    \centering
     \setlength{\tabcolsep}{18.0pt}
    \caption{ Quantitative comparison with SOTA methods across five widely-used datasets, \textit{i.e.}, Cotton80, SoyLoc, SoyGene, SoyAgeing, and SoyGlobal. Here, R50 denotes the ResNet-50 model \cite{He2015Deep}, ViT-B represents the Vision Transformer base model \cite{DBLP:conf/iclr/DosovitskiyB0WZ21}, and Swin-B corresponds to the Swin Transformer base model \cite{DBLP:conf/iccv/LiuL00W0LG21}.}
    \begin{tabular}{l c c c ccc}
    	\toprule[1pt]
       \multirow{2}{*}{Method} & \multirow{2}{*}{Backbone}&  \multicolumn{5}{c}{Top 1 Accuracy (\%)} \\
       \cline{3-7}
       &&Cotton80 & SoyLoc &SoyGene & SoyAgeing & SoyGlobal  \\
       \toprule[0.7pt]
       FDCL-DA  \cite{DBLP:journals/pr/ChenJWDLWW24} &    ResNet-50 & 43.3&49.8&70.0&76.9&54.2\\
        ADL  \cite{DBLP:journals/pami/ChoeLS21} &  ResNet-50 & 43.8&34.7&55.2&61.7&39.4\\
        Cutmix \cite{DBLP:conf/iccv/YunHCOYC19} &  ResNet-50 & 45.0&26.3&66.4&62.7&30.3 \\
       SimCLR \cite{DBLP:conf/icml/ChenK0H20} & ResNet-50 & 51.7 & 37.3 & 62.7 & 64.7 & 42.5 \\
       BYOL  \cite{DBLP:conf/nips/GrillSATRBDPGAP20}  & ResNet-50 & 52.9 &33.2&60.7&64.8&41.4 \\
       DCL  \cite{DBLP:conf/cvpr/ChenBZM19} &  ResNet-50 & 53.8&45.3&71.4&73.2&42.2\\
       MaskCOV  \cite{DBLP:journals/pr/YuZGX21} &  ResNet-50 & 58.8&46.2&73.6&75.9&50.3 \\
       CSDNet \cite{DBLP:journals/tcsv/FangJTL24}  &  ResNet-50 & 61.7&48.2&66.5&78.0&51.1 \\
       DeiT \cite{DBLP:conf/icml/TouvronCDMSJ21} & ViT-B & 54.2&38.7&66.8&69.5&45.3 \\
       TransFG  \cite{DBLP:conf/aaai/HeCLKYBW22} & ViT-B & 54.6&40.7&22.4&72.2&21.2 \\
       SIM-OFE  \cite{DBLP:journals/tip/SunHXP24} & ViT-B & 54.6&25.0&15.5&34.8&70.7 \\
       ILA \cite{DBLP:journals/corr/abs-2409-11051}& ViT-B & 55.4& 50.8&62.2 & 75.0 & 58.1 \\
       Mix-ViT \cite{DBLP:journals/pr/YuWZG23} & ViT-B\&R50 & 60.4&56.2&79.9&76.3&51.0 \\
       CLE-ViT \cite{DBLP:conf/ijcai/0001WG23} & Swin-B & 63.3&47.2&78.5&82.1&75.2 \\
       CSDNet \cite{DBLP:journals/tcsv/FangJTL24} & Swin-B & 67.9&60.5&86.9&83.2&76.2\\
        \toprule[0.7pt]
 \rowcolor{gray!30}
 Our GAEor & Swin-B & \textbf{71.3}&\textbf{62.5}&\textbf{88.7
}&\textbf{86.4}&\textbf{81.2}\\
		\toprule[1pt]

    \end{tabular}
    
   \label{tab:com1}
\end{table*}

\begin{table*}[!t]
    \centering
     \setlength{\tabcolsep}{18.5pt}
    \caption{ The classification accuracy comparisons  on the five subsets of the SoyAgeing dataset, namely R1, R3, R4, R5, and R6. Here, "Average" represents the average precision across the five subsets. R50 denotes the ResNet-50 model \cite{He2015Deep}, ViT-B represents the Vision Transformer base model \cite{DBLP:conf/iclr/DosovitskiyB0WZ21}, and Swin-B corresponds to the Swin Transformer base model \cite{DBLP:conf/iccv/LiuL00W0LG21}.
    }
    \begin{tabular}{l c c c c ccc}
    	\toprule[1pt]
       \multirow{2}{*}{Method} & \multirow{2}{*}{Backbone}&  \multicolumn{6}{c}{Top 1 Accuracy (\%)} \\
       \cline{3-8}
       &&R1&R3&R4&R5&R6&Average\\
       \toprule[0.7pt]
       SimCLR  \cite{DBLP:conf/icml/ChenK0H20} & ResNet-50 & 53.6&45.7&45.4&50.4&35.9&46.2 \\
        ADL  \cite{DBLP:journals/pami/ChoeLS21} &  ResNet-50 & 66.7&58.9&64.8&68.5&49.7&61.7\\
        Cutmix  \cite{DBLP:conf/iccv/YunHCOYC19}&  ResNet-50 & 65.6&59.2&64.2&68.8&53.6&62.3 \\
       BYOL \cite{DBLP:conf/nips/GrillSATRBDPGAP20}  & ResNet-50 & 71.1&66.2&66.2&64.7&56.1&64.8 \\
       DCL \cite{DBLP:conf/cvpr/ChenBZM19} &  ResNet-50 & 76.9&73.8&76.2&76.2&62.9&73.2\\
       MaskCOV  \cite{DBLP:journals/pr/YuZGX21} &  ResNet-50 & 79.8&74.7&79.6&78.3&67.0&75.9 \\
       FDCL-DA \cite{DBLP:journals/pr/ChenJWDLWW24} &    ResNet-50 & 76.4&76.2&79.1&82.8&70.1&76.9\\
       
       DeiT \cite{DBLP:conf/icml/TouvronCDMSJ21} & ViT-B & 73.0&70.4&69.1&74.7&60.5&69.5 \\
       SIM-OFE \cite{DBLP:journals/tip/SunHXP24} & ViT-B & 69.9&73.2&73.1&73.9&63.2&70.7 \\
       TransFG  \cite{DBLP:conf/aaai/HeCLKYBW22} & ViT-B & 75.0&74.6&74.2&76.2&60.8&72.2 \\
       Mix-ViT  \cite{DBLP:journals/pr/YuWZG23} & ViT-B\&R50 & 79.3&77.2&78.0&79.2&67.9&76.3 \\
       CLE-ViT \cite{DBLP:conf/ijcai/0001WG23} & Swin-B & 80.8&83.3&84.2&86.4&76.0&82.1 \\
       CSDNet \cite{DBLP:journals/tcsv/FangJTL24} & Swin-B & 83.8&85.2&85.2&84.9&76.9&83.2\\
        \toprule[0.7pt]
 \rowcolor{gray!30}
 Our GAEor & Swin-B & \textbf{85.7}&\textbf{88.8}&\textbf{90.2}&\textbf{88.9}&\textbf{77.9}&\textbf{86.4}\\

       \toprule[1pt]
       \end{tabular}
       \label{tab:com2}
       \end{table*}

\subsection{Ablation Experiments}
\textbf{Effect of various components.}
By individually adding the major components of GAEor and conducting an ablation study as shown in Tab.~\ref{tab:ae}, we validate the effectiveness of these components on two datasets (Cotton80 and SoyLoc) and reveal their critical contributions in distinguishing highly similar objects under limited data.
Using Swin Transformer  \cite{DBLP:conf/iccv/LiuL00W0LG21} as the backbone achieves baseline accuracies of 61.3\% and 51.8\% on Cotton80 and SoyLoc, respectively, validating its capacity to capture subtle discriminative features. 
The SDA module enhances fine-grained perception, leading to accuracy improvements of 1.9\% and 3.0\% on Cotton80 and SoyLoc, respectively, underscoring its effectiveness in amplifying subtle yet critical visual discrepancies. The GAE module, which focuses on extracting geometric attributes, yields even greater gains of 3.0\% and 3.9\%, highlighting the discriminative power of structural patterns in distinguishing highly similar objects.
While the standalone impact of the GAT module is not explicitly quantified, its integration facilitates the fusion of geometric attributes with visual cues, forming a hybrid representation that bridges localized detail and global structural consistency. Notably, the synergistic combination of all three modules results in substantial performance improvements (10.0\% and 11.7\% on Cotton80 and SoyLoc, respectively), demonstrating the complementary learning mechanisms: SDA refines local discriminative details, GAE encodes geometric invariants, and GAT ensures their cohesive integration into the classification pipeline.
This hierarchy of improvements emphasizes that addressing both visual and structural distinctiveness is essential for robust recognition in data-scarce, high-similarity scenarios. 

\textbf{Importance of different constraints.}
The ablation study underscores the pivotal role of each loss function in guiding GAEor to explore alternative discriminative cues for distinguishing highly similar objects, as demonstrated in Tab.~\ref{tab:ael}. When solely relying on the classification loss ($\mathcal{L}_{\mathrm{cls}}$), the model achieves marginal performance gains, indicating its inadequacy in capturing decision-making cues for highly similar objects . 
Introducing the regularization term ($\mathcal{L}_{\mathrm{reg}}$) suppresses irrelevant visual noise but offers limited discriminative improvement, suggesting that mere feature suppression is insufficient for resolving high-similarity ambiguities. 
A critical leap occurs with the integration of the polar distance constraint ($\mathcal{L}_{\mathrm{dis}}$) and geometric attribute transfer loss ($\mathcal{L}_{\mathrm{tat}}$), which collectively enforce relational consistency among discrepancies—distance constraints establish metric relationships, while $\mathcal{L}_{\mathrm{tat}}$ aligns structural attributes with classification objectives. 
This synergy achieves substantial accuracy boosts by explicitly linking geometric attributes to discriminative learning. Further augmenting these with the polar angle constraint ($\mathcal{L}_{\mathrm{ang}}$) refines angular relationships among features, ensuring a geometrically coherent embedding space that captures both radial and angular distinctiveness. The combined effect of $\mathcal{L}_{\mathrm{tat}}$, $\mathcal{L}_{\mathrm{dis}}$, and $\mathcal{L}_{\mathrm{ang}}$ demonstrates that multi-faceted geometric constraints—spanning distance, angle, and attribute transfer—are indispensable for discovering alternative discriminative cues. 
These losses collectively transform the model’s focus from isolated visual cues to holistic geometric reasoning, where geometric priors mitigate challenge in data-scarce scenarios. The improvements reflects that each constraint addresses unique aspects of geometric representation, and their omission disrupts the delicate balance required for robust geometric attribute modeling.

\subsection{Comparison with State-of-the-art Methods}
We commence by contrasting the quality of our proposed GAEor with that of previous Ultra-FGVC methods. Tab.~\ref{tab:com1} showcases the performance of various competitive methods across five benchmarks, namely Cotton80, SoyLoc, SoyGene, SoyAgeing, and SoyGlobal.
Compared to existing works, which primarily focus on increasing sample diversity and capturing hard-to-identify discriminative discrepancies, our GAEor distinguishes itself by equipping the classification network with the ability to model and parameterize geometric attributes, thereby uncovering novel alternative recognition cues.
Therefore, GAEor consistently demonstrates superior recognition performance compared to other methods, including CLE-ViT \cite{DBLP:conf/ijcai/0001WG23} and FDCL-DA \cite{DBLP:journals/pr/ChenJWDLWW24}, across all tested backbone networks. 
This highlights that investigating the intrinsic geometrical features of highly similar objects can indeed uncover more discriminative cues compared to solely exploring pixel-level cues.
CLE-ViT \cite{DBLP:conf/ijcai/0001WG23} employs a data augmentation strategy to generate more diverse samples and leverages contrastive learning to capture discriminative cues, achieving encouraging results across the five benchmarks.  
Hence, GAEor first modulates the discrepancies to enhance their perceptibility and facilitates the backbone network in predicting the relative polar coordinates of these discrepancies, effectively exploring unique geometric attributes.  
Ultimately, GAEor achieves outstanding results on the Cotton80 and SoyGlobal datasets, surpassing the recent state-of-the-art CSDNet \cite{DBLP:journals/tcsv/FangJTL24} by 3.4\% and 5.0\%, respectively.

\subsection{Comparison Across Soybean Growth Stages }
The SoyAgeing dataset presents a distinctive challenge in Ultra-FGVC by organizing images into five subsets corresponding to distinct cultivation stages, which amplifies intra-class variations caused by morphological and physiological changes during growth. 
The comparative results for all competing methods across these five subsets are presented in Tab.~\ref{tab:com2}.
Traditional methods such as CSDNet \cite{DBLP:journals/tcsv/FangJTL24}, which emphasize local discriminative discrepancies, often struggle to maintain robustness across different cultivation stages due to their sensitivity to appearance shifts. We hypothesize that this limitation arises from the instability of pixel-level cues (e.g., leaf color), which can vary significantly across growth stages. In contrast, our GAEor framework captures invariant geometric associations intrinsic to soybean cultivars, reducing dependence on such fluctuating visual features. Notably, GAEor achieves a 3.2\% improvement in average accuracy over the state-of-the-art CSDNet, demonstrating the advantage of geometric attribute learning. These results highlight GAEor’s superior capability in addressing the challenges of Ultra-FGVC, especially under settings with substantial intra-class variation.

\begin{table}[t]
    \centering
    \setlength{\tabcolsep}{18pt}
    \caption{Evaluation of diverse visual saliency feedback, \textit{i.e.,} the classification activation map (CAM) and the geometry-relevant visual feedback (SVF) generated by our SDA module (\S\ref{sec3.2}), on the Cotton80 and SoyLoc datasets. }
    \begin{tabular}{ccc}
     \toprule[1pt]
     Method & Cotton80 & SoyLoc  \\
        \toprule[0.7pt]
      GAEor w CAM  &  66.7\% & 57.9\% \\
      \rowcolor{gray!30}
     GAEor W SVF &  \textbf{71.3\%} & \textbf{62.5\%} \\
         \toprule[1pt]
    \end{tabular}
    \label{tab:feedback}
\end{table}

\begin{table}[t]
    \centering
    \setlength{\tabcolsep}{9pt}
    \caption{Evaluation results on the Cotton80 and SoyLoc datasets for establishing geometric attributes without and with the guidance of geometry-relevant details ($\mathbf{1_{(\mathbf{M}{(\mathrm{m,n})} > \mu_{\mathbf{M}})}}$) in $\mathcal{L}_{\mathrm{dis}}$ and $\mathcal{L}_{\mathrm{ang}}$. }
    \begin{tabular}{l c c}
    \toprule[1pt]
      Loss function   &  Cotton80 & SoyLoc \\
       \toprule[0.7pt]
        $\mathcal{L}_{\mathrm{dis}}$ \& $\mathcal{L}_{\mathrm{ang}}$ w/o $\mathbf{1_{(\mathbf{M}{(\mathrm{m,n})} > \mu_{\mathbf{M}})}}$  & 69.7\% &59.3\% \\
       \rowcolor{gray!30}
        $\mathcal{L}_{\mathrm{dis}}$ \& $\mathcal{L}_{\mathrm{ang}}$ w $\mathbf{1_{(\mathbf{M}{(\mathrm{m,n})} > \mu_{\mathbf{M}})}}$  & \textbf{71.3\%} & \textbf{62.5\%} \\
         \toprule[1pt]
    \end{tabular}
    \label{tab:DF}
\end{table}

\begin{table}[t]
    \centering
    \setlength{\tabcolsep}{18pt}
    \caption{Evaluation results on the Cotton80 and SoyLoc datasets, emphasizing invariant characteristics in learning geometric attributes. Here, SD denotes the standard deviation operation. }
    \begin{tabular}{ccc}
     \toprule[1pt]
     Loss function & Cotton80 & SoyLoc  \\
        \toprule[0.7pt]
      $\mathcal{L}_{\mathrm{ang}}$ w/o SD  &  65.3\% & 56.2\% \\
      \rowcolor{gray!30}
     $ \mathcal{L}_{\mathrm{ang}}$ w SD  &  \textbf{71.3\%} & \textbf{62.5\%} \\
         \toprule[1pt]
    \end{tabular}
    \label{tab:SD}
\end{table}

\begin{table}[t]
    \centering
    \setlength{\tabcolsep}{16pt}
    \caption{Evaluation results on the Cotton80 and SoyLoc datasets for establishing geometric associations of geometry-relevant details using Cartesian or polar coordinates, respectively. }
    \begin{tabular}{ccc}
     \toprule[1pt]
     Type & Cotton80 & SoyLoc  \\
        \toprule[0.7pt]
       Cartesian Coordinates &  60.1\% & 51.6\% \\
      \rowcolor{gray!30}
     Polar Coordinates  &  \textbf{71.3\%} & \textbf{62.5\%} \\
         \toprule[1pt]
    \end{tabular}
    \label{tab:trans}
\end{table}

\subsection{Further Analysis}
\textbf{Impact of diverse visual feedback in SDA (\S \ref{sec3.2}).}
To validate the effectiveness of diverse visual feedback mechanisms in geometric attribute modeling, we compared saliency-guided feedback with the commonly used Classification Activation Map (CAM) in Tab.~\ref{tab:feedback}. While CAM is effective in highlighting category-discriminative regions, it inherently emphasizes class-specific patterns (e.g., color or texture) over spatially structured relationships. This class-prioritized attention leads to a misalignment with the goals of geometric modeling, as CAM lacks explicit mechanisms for enhancing geometrically coherent details. Consequently, it struggles to capture the structural regularities required for robust geometric attribute learning. In contrast, GAEor’s lightweight generator—trained in an end-to-end manner—dynamically adapts visual representations to enhance structurally salient regions. By reinforcing spatially adaptive feedback, it functions analogously to attention mechanisms that preserve geometric consistency across varied object instances. The amplified regions identified by the generator help ease the challenge of predicting geometric attributes, promoting a more stable and geometry-aware representation.

\textbf{Effect of the guidance of geometry-relevant details.} 
To evaluate the impact of geometry-relevant detail guidance on modeling geometric attributes, we conducted an ablation study by removing its constraints from polar distance (Eq.\ref{eq9}) and polar angle (Eq.\ref{eq10}) computations. As shown in Tab.~\ref{tab:DF}, the absence of this guidance forces GAEor to rely on holistic image features, aggregating all visual cues indiscriminately for geometric attribute calculation. This approach introduces noise from irrelevant regions, diluting critical structural relationships and leading to a performance decline. 
In contrast, integrating geometry-relevant detail guidance enables the network to prioritize spatially adaptive structural features, which focus on gradient maps and long-range dependencies to suppress noise in low-visibility regions. By aligning geometric modeling with explicit structural priors—the network effectively pays more attention to geometry-sensitive details while ignoring redundant information. This hard attention enhances recognition of highly similar samples by explicitly emphasizing subtle structural distinctions, ultimately improving robustness and accuracy. 


\textbf{Exploration of rotation-invariant property.}
To investigate the necessity of rotation-invariant modeling for geometric attributes, we analyzed two strategies for constraining polar angle deviations  (Eq.~\ref{eq10}) between predictions and ground truths. 
As shown in Tab.~\ref{tab:SD}, directly minimizing raw angular differences fails to enforce rotation invariance because the unconstrained optimization allows arbitrary angular shifts, causing inconsistent geometric representations across different objects with the same category. 
This instability degrades recognition performance, particularly for highly similar objects with diverse orientation variations in real-world scenarios. 
Conversely, minimizing the standard deviation of angular differences imposes statistical stability on the optimization process, forcing the network to focus on orientation-agnostic geometric patterns rather than transient angular offsets. This constraint aligns with the intrinsic property of rotation invariance by penalizing dispersion in angular deviations, thereby consolidating consistent structural relationships regardless of object rotations. The significant accuracy improvement confirms that explicit statistical regularization (standard deviation) is essential for GAEor to preserve rotation-invariant geometric attributes, which fundamentally enhances robustness against viewpoint variations while maintaining discriminative power.

\textbf{Necessity of the Cartesian-to-polar coordinate transformation.}
Tab.~\ref{tab:trans} illustrates the performance evolution when transitioning from Cartesian to polar coordinates for establishing geometric associations of details within objects. When using Cartesian coordinates, GAEor directly predicts the spatial coordinates of each detail within the object. However, this approach struggles to establish geometric associations and introduces noise, leading to a significant performance drop. In contrast, by using polar coordinates, GAEor better handles geometric variations, ensuring that geometry-relevant features are consistently recognized across different orientations and scales. Converting pixel locations into polar coordinates enables the model to encode relative positional information more robustly, which is crucial for tasks that require high sensitivity to subtle structural details.

\begin{figure}[!t]
\begin{center}
   \includegraphics[width=1.\linewidth]{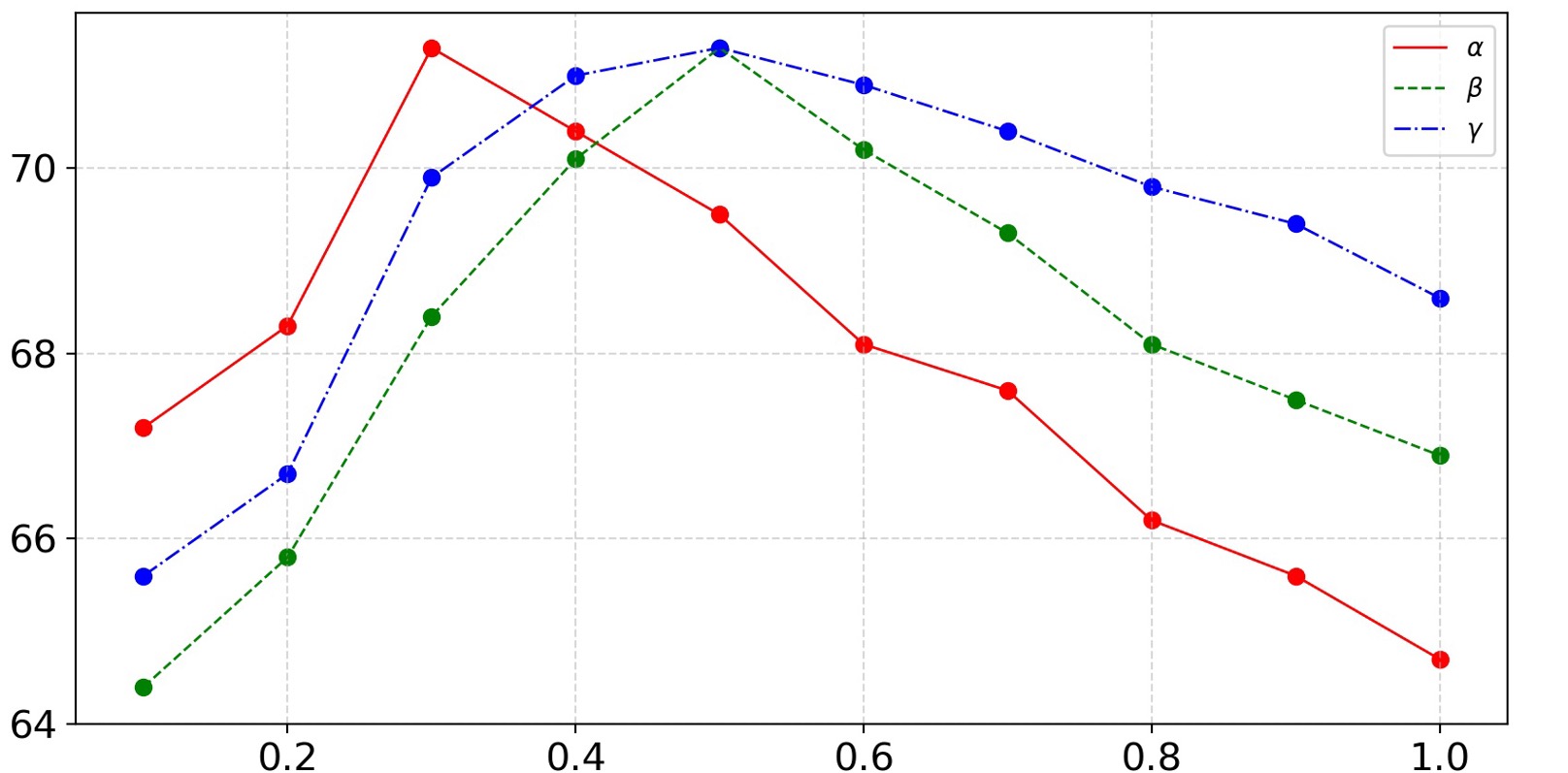}
   \caption{Analyses of hyper-parameters $\alpha$, $\beta$ and $\gamma$ in Eq.~\ref{eq13}. The results denote Top-1 Accuracy on Cotton80.}
   \label{fig:loss}
\end{center}
\end{figure}

\textbf{Hyperparameter Analyses.} The sensitivity analysis of hyperparameters \(\alpha\), \(\beta\), and \(\gamma\) in Eq.~\ref{eq13} reveals distinct roles and impacts on GAEor’s performance (Fig.~\ref{fig:loss}). Specifically, \(\alpha\) governs the balance of geometry-relevant details by modulating their response range, \(\beta\) regulates the self-supervised loss term for geometric property learning, and \(\gamma\) controls the supervised loss weight for transferring geometric attributes to the classification branch. Experimental results demonstrate that GAEor’s performance fluctuates with variations in these parameters, confirming their non-negligible influence on the model. For instance, deviations from the optimal values (\(\alpha=0.3\), \(\beta=0.5\), \(\gamma=0.5\)) lead to suboptimal outcomes, suggesting that \(\alpha\) requires finer balancing compared to \(\beta\) and \(\gamma\), which share identical optimal weights. These findings highlight the necessity of carefully calibrating each hyperparameter to harmonize discriminative learning, geometric learning, and attribute-classification alignment in the proposed framework.

\begin{figure}[!t]
\begin{center}
   \includegraphics[width=1.\linewidth]{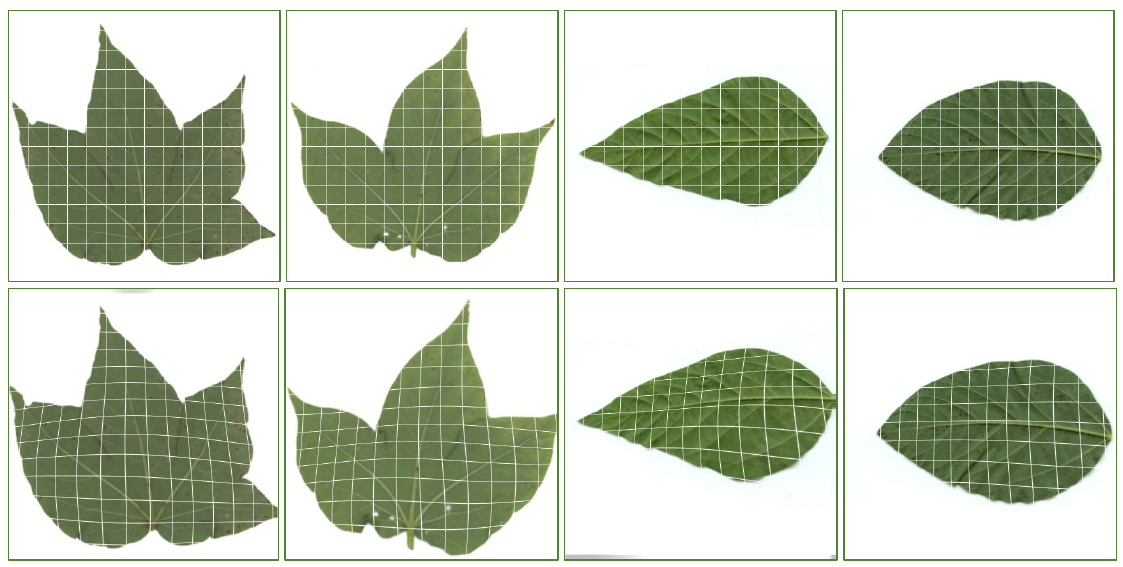}
   \caption{Visualization of the effect of feedback-driven discrepancy modulation module. The first and second rows display the original and transformed inputs, respectively. 
   }
   \label{fig:zoom}
\end{center}
\end{figure}

\begin{figure}[t]
\begin{center}
   \includegraphics[width=1.\linewidth]{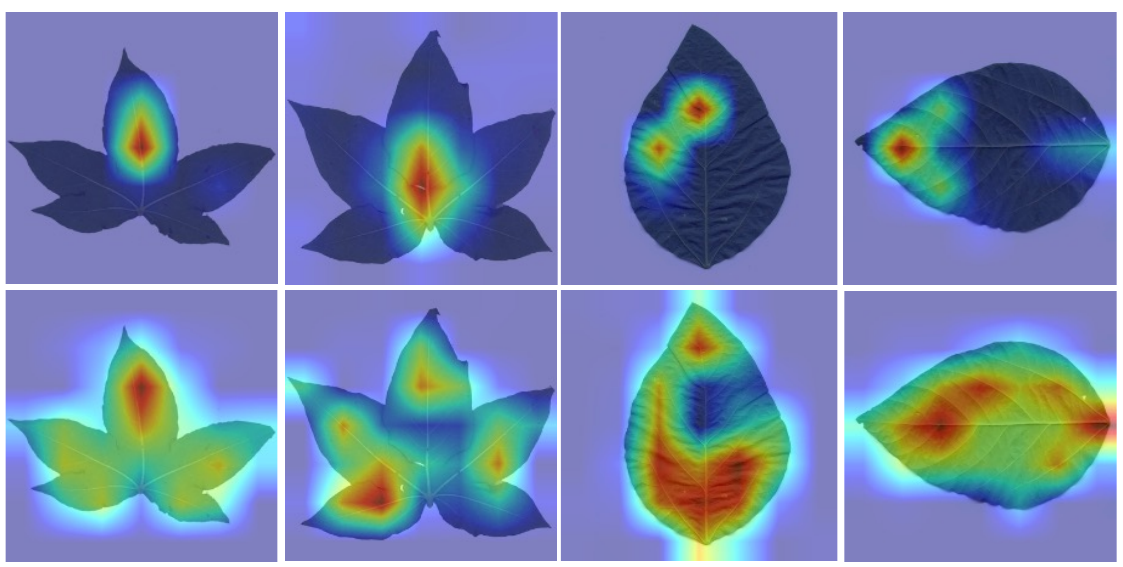}
   \caption{Visualization of the impact of geometric attribute parameterization. The first and second rows are CAMs using the backbone network and our GAEor, respectively. 
   }
   \label{fig:activation}
\end{center}
\end{figure}

\subsection{Geometric Attribute Analysis} 
Interpreting geometric attributes remains challenging, as these attributes are optimized within a latent space, making direct analysis difficult. To address this, 
We adopt an indirect approach by visualizing the sources of geometric attributes (Fig.~\ref{fig:zoom}), which reveals the underlying content, and the corresponding features influenced by these attributes (Fig.~\ref{fig:activation}), enabling us to indirectly trace and understand the role of geometric attributes.

As shown in Fig.~\ref{fig:zoom}, these visualizations demonstrate that subtle visual cues within local regions are effectively enhanced through inhomogeneous transformations while suppressing background noise and non-critical features. Note that grid lines are manually overlaid on the images to clearly highlight pixel shifts in the transformed inputs.
Importantly, although the task-relevant details are zoomed, their structural integrity remains largely intact, with only minor perturbations. These amplified vital details in the transformed images become more accessible to the backbone network, facilitating the extraction of geometric attributes.

Additionally, we provide visualizations to illustrate the impact of geometric attributes. The comparative visualizations of the baseline and our proposed model are presented in Fig.~\ref{fig:activation}. These results demonstrate that our model attends to more extensive visual cues rather than the fixed regions predicted by the baseline. Notably, the first two columns and the last two columns correspond to sub-figures from the same categories, respectively. These highlighted regions in the second row roughly align with certain attributes of highly similar objects, such as the five corners of a cotton leaf or the tip and tail of a soybean leaf.
The results indicate that the activation of object parts is evidently related to category-specific attributes, offering a clear explanation for the success in identifying highly similar objects.

\begin{table}\centering
		\caption{ Comparison of different methods on CUB-200-2011 and Stanford Cars datasets.} 
		
\setlength{\tabcolsep}{12pt}
	\begin{tabular}{c|c|c}
		\toprule[1pt]
		Method& CUB-200-2011 & Stanford Cars  \\
		\toprule[0.7pt]
		
		SCDA \cite{Wei2016Selective} & $ 80.1\% $ & 92.8\% \\
		AutoBD \cite{Yao2018AutoBD}  & $ 81.6\% $ &88.9\%  \\
		OPAM \cite{Peng2017Object} &   85.8\%  & 92.2\%\\
		Kernel-Activation \cite{DBLP:conf/iccv/CaiZZ17}&  $ 85.3\% $ & 91.7\% \\
		Kernel-Pooling \cite{DBLP:conf/cvpr/CuiZWLLB17} & 86.2\% & 92.4\% \\
        DBT-Net \cite{DBLP:conf/nips/ZhengFZL19}& 88.1\% & 94.5\% \\	
        GaRD \cite{DBLP:conf/cvpr/ZhaoYHL21} &  89.6\%  & 95.1\%  \\	
ACNet \cite{DBLP:conf/cvpr/JiWZDWZLH20} &  88.1\%  & 94.6\%\\

S3Ns \cite{Ding_2019_ICCV} &   88.5\%  & 94.7\%	\\
SPS \cite{DBLP:conf/iccv/HuangWT21} &  88.7\% & 94.9\% \\
P2P-Net \cite{DBLP:conf/cvpr/YangWCX022} &  90.2\% & 95.4\%	\\			
CAL \cite{DBLP:conf/iccv/Rao0L021}  &  90.6\% & 95.5\% \\
 SRGN  \cite{DBLP:journals/ijcv/WangWLCOT24}&91.4\% & 95.8\%\\
 \toprule[0.7pt]
 \rowcolor{gray!30}
  Our GAEor   &\textbf{92.2\%} & \textbf{96.0\%}\\
		\toprule[1pt]
	\end{tabular}
    \label{fgvc}

\end{table}

\subsection{Geometric Attribute Extension}
To validate the generalizability of geometric attributes as robust discriminative cues, we evaluate GAEor on the CUB-200-2011 and Stanford Cars benchmarks under real-world scenarios. 
Existing approaches, such as SRGN~\cite{DBLP:journals/ijcv/WangWLCOT24} and CAL~\cite{DBLP:conf/iccv/Rao0L021}, primarily focus on pixel-level discriminative cues to distinguish visually similar objects, achieving strong recognition performance.
In contrast, our method, GAEor, investigates an alternative source of discriminative information—geometric attributes—which provide a complementary perspective beyond raw pixel-level representations.
As evidenced in Tab.\ref{fgvc}, GAEor effectively recognizes visually similar categories even in complex natural environments—achieving consistent performance gains over SRGN (e.g., +0.8\% on cross-domain CUB-200-2011 dataset). This success extends beyond plant-specific applications (e.g., soybean cultivars in Fig.~\ref{fig:introduction1a}(a) to diverse real-world objects.

\section{Conclusion}
In this paper, we propose GAEor, a novel framework for investigating the geometrical features of highly similar objects, aiming to uncover novel alternative recognition cues for Ultra-FGVC.
GAEor first amplifies vital details via gradient-guided visual saliency, then encodes their spatial associations using polar coordinate-based geometric embeddings into the final representation.
By translating the spatial relationships of vital details into geometric attributes, the network establishes structural descriptors that complement conventional pixel-level visual cues, thus uncovering novel alternative recognition cues. 
Extensive experiments demonstrate that GAEor significantly outperforms state-of-the-art methods, highlighting the effectiveness of geometric attribute modeling in distinguishing highly similar objects.



\ifCLASSOPTIONcaptionsoff
  \newpage
\fi

\bibliographystyle{IEEEtran}
\bibliography{egbib}

\end{document}